# Application of Artificial Neural Networks in Aircraft Maintenance, Repair and Overhaul Solutions

Soumitra Paul[1], Kunal Kapoor[2], Devashish Jasani[3], Rachit Dudhwewala[4], Vijay Bore Gowda[5] , T.R.Gopalakrishnan Nair[6]

[1] 3rd Semester, B.E. IT, DSI, Bangalore. soumitra@aeroit.co.in
[2] 3rd Semester, B.E. CSE, DSI,Bangalore, kunalkapoor.com@gmail.com
[3] 3rd Semester, B.E. ISE, DSI, Bangalore, devashish_here@hotmail.com
[4] 3rd Semester, B.E. Telecom, DSI, Bangalore, rachitdudhwewala@yahoo.com
[5] Project Manager, Incubation Centre, Research Industry Incubation Centre, DSI, Bangalore, vijay.b@aeroit.co.in

**ABSTRACT**
This paper reviews application of Artificial Neural Networks in Aircraft Maintenance, Repair and Overhaul (MRO). MRO solutions are designed to facilitate the authoring and delivery of maintenance and repair information to the line maintenance technicians who need to improve aircraft repair turn around time, optimize the efficiency and consistency of fleet maintenance and ensure regulatory compliance. The technical complexity of aircraft systems, especially in avionics, has increased to the point at which it poses a significant troubleshooting and repair challenge for MRO personnel. As per the existing scenario, the MRO systems in place are inefficient. In this paper, we propose the centralization and integration of the MRO database to increase its efficiency. Moreover the implementation of Artificial Neural Networks in this system can rid the system of many of its deficiencies. In order to make the system more efficient we propose to integrate all the modules so as to reduce the efficacy of repair.

## INTRODUCTION

Last few years have been turbulent for the aviation industry owing to the unprecedented rise in its commodities due to inflation. It gets very challenging for the industry now to keep its costs as low as possible and generate maximum revenue. For this to happen the industry needs to ensure that its asset utilization is optimum. Therefore the maintenance management system of the existing aircrafts needs to be spot on to ensure that they spend maximum time in air so as to make the best use of its machinery. Hence the Maintenance, Repair and Overhaul (MRO) services need to be very efficient so that the aircraft's 'hangar- time' for MRO purposes is as low as possible.

MRO implementation in the aviation industry implies the improvement of processes from the hanger to the flying hours of the aircraft by creating a framework for its total asset management. MRO is responsible for providing on site technical support and engineering disposition in order to reduce aircraft problems.

It also acts as a technical liaison with aircraft engine and component manufacturers about overhaul, repair and its modifications as required by aircraft policies, maintaining the system performance and recommending action to correct abnormal trends to improve the engine reliability. It also develops engineering modifications which retrofits to the fleet in accordance with OEM service bulletins. The complex business of aircraft maintenance, repair and overhaul makes extremely high demands on quality, security and accountability. IT solutions for this area are focussing increasingly on supporting these highly sensitive processes efficiently and cost-effectively. This system though has its flaws, it is inefficient and there is a vast scope of improvement in this field. The Original Equipment Manufacturer (OEM) manual is an encyclopedia on the product. It contains all the information about the product starting from its components, features, troubleshooting, repair and its lifespan. Presently different products have different OEM manuals and if this entire data is centralized then each products database will support the other thereby filtering the data and increasing the probability of decision making and support systems.

Application of Artificial Neural Networks (ANN) in the MRO will help us get rid of many of its superfluous data and help us estimate probability at the point and the extent of damage caused in an aircraft. Neural Networks can help better detect and estimate aircraft unit fault diagnosis, corrosion detection in ageing materials, real time assessment of engine conditions and source of damage to unconventional structures. The application of this ANN in the MRO will help the system to work more efficiently and if all the different modules are integrated into a single computed database then that will further increase the accuracy of the system.

## CURRENT MRO SYSTEM

Fig. 1 shows the block diagram of the current MRO system. As per the existing system, there is a time to time scheduled checking of the aircraft. Depending upon the inspection carried out faults are detected. As of now, each product of an aircraft comes with an OEM manual which is provided by the product manufacturer. The OEM manual contains detailed information about product, its component and types and its related issues and its proper diagnosis.

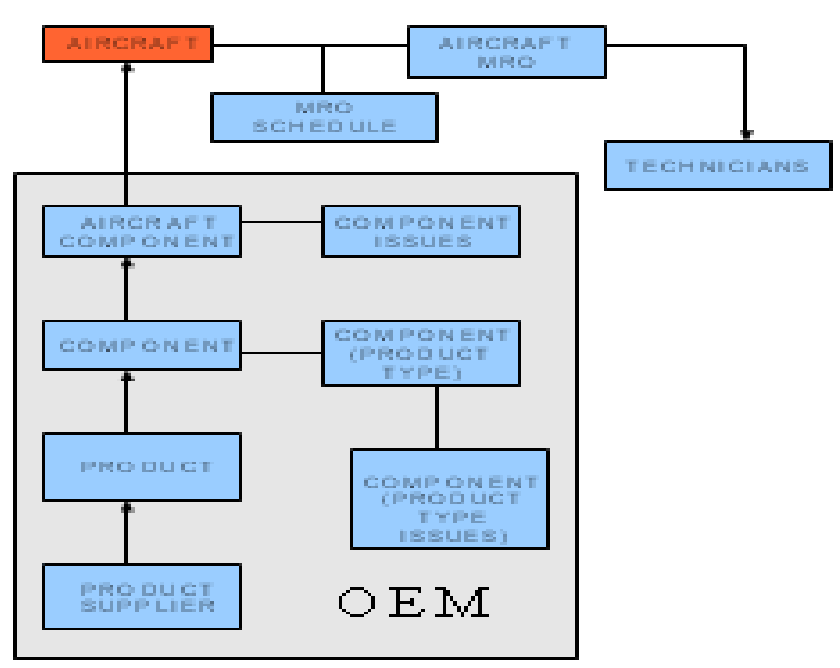

Figure 1: Block Diagram of the Current System

This OEM manual needs to be updated time to time as per the instructions issued by the regulatory body. Based on this data the MRO assesses the information and gives the input to the technicians regarding fault assessment.[8] The technicians then act accordingly as per the inputs given by the MROs to them. This system though has its own flaws. It is highly inefficient. There is no real time assessment of the data.

As clearly shown in Fig.2, a radical amount of data is either incorrect, hard to find or incomplete.[2]In order to increase its efficiency we propose the centralization of the data given by the OEM manual and the integration of each of the parts. This method will have better probability of finding the problem node in the aircraft and it's most suited and immediate repair. Moreover, we require a system which can give the near to accurate probability of any fault at any point in an aircraft and its most suitable, cost effective and viable diagnosis.

One of the methods by which this can be made possible is by introducing the concept of Artificial Neural Networks (ANN) to the system.

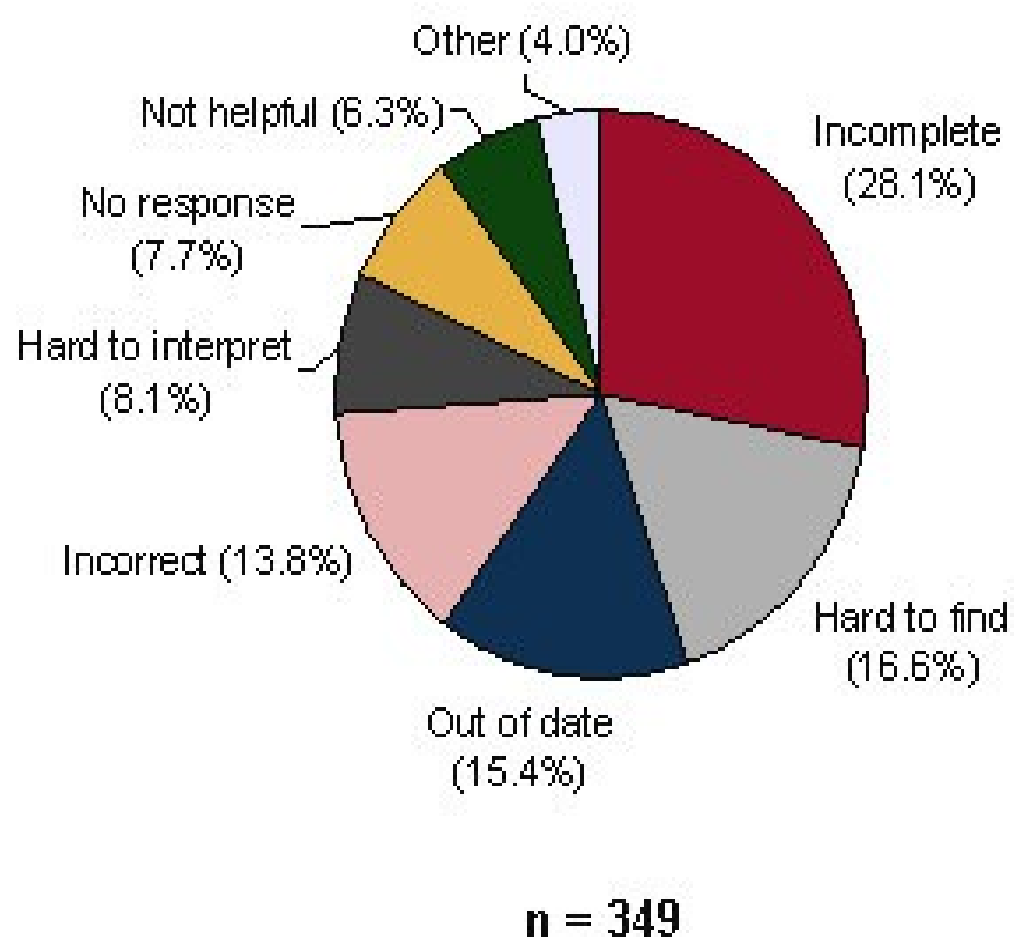

**Figure 2: Perceived Problems in Aircraft Maintenance Information**

## NEURAL NETWORKS

An Artificial Neural Network (ANN) is a mathematical model or computational model based on biological neural networks. It consists of an interconnected group of artificial neurons and processes information using a connectionist approach to computation.[5] In more practical terms neural networks are non- linear statistical data modeling tools. They can be used to model complex relationships between inputs and outputs or to find patterns in data. Since neural networks are best at identifying patterns or trends in data, they are well suited for prediction or forecasting.

## SELF ORGANISING MAPS (SOM)

The self-organizing map (SOM) is an excellent tool in exploratory phase of data mining. It projects input space on prototypes of a low-dimensional regular grid that can be effectively utilized to visualize and explore properties of the data. When the number of SOM units is large, to facilitate quantitative analysis of the map and the data, similar units need to be grouped, i.e., clustered [1]. Self- organizing map (SOM) is a type of artificial neural network that is trained using unsupervised learning toproduce a low-dimensional (typically two dimensional), discrete representation of the input space of the training samples, called a map. The map seeks to preserve the topological properties of the input space [11].

This makes SOM useful for visualizing low-dimensional views of high- dimensional data, akin to multidimensional scaling.

**MULTI-DIMENSIONAL SCALING**

It is a set of related statistical techniques often used in information visualization for exploring similarities or dissimilarities in data. MDS is a special case of ordination. An MDS algorithm starts with a matrix of similar items, then assigns a location of each item in a low-dimensional space, suitable for graphing or 3D visualization [6].

**FUZZY LOGIC**

Fuzzy logic is a form of multi-valued logic derived from fuzzy set theory to deal with reasoning that is approximate rather than precise. Just as in fuzzy set theory the set membership values can range (inclusively) between 0 to 1, in fuzzy logic the degree of truth of a statement can range between 0 to 1 and is not constrained to the two truth values i.e. true or false as in classic predicate logic. And when *linguistic variables* are used, these degrees may be managed by specific functions [3].

Fuzzy logic is a new way of expressing probability. While both fuzzy logic and probability theory can be used to represent subjective belief, fuzzy set theory uses the concept of fuzzy set membership. The fuzzy probability can also be generalized as a possible model.

**APPLICATION OF NEURAL NETWORKS WORK IN MRO**

Neural network has a great role in aircraft fault diagnosis. It uses historical data stored to analyze the condition and trace the fault. However, neural networks cannot be the deciding factor as it is based on probability. Hence it helps man to decide the problem and work upon it by giving the probability of damage.

The system works according to the following procedure:

1. The information regarding the aircraft is fed into the neural system.
2. The system analyzes the data and checks for historical models corresponding to the data.
3. Then based on the records, the system throws out probabilities.
4. An experimental model is compared to the result of the ANN model.

This neural system gives man a better idea of the current problem or condition and hence increases efficiency of the whole MRO system.

This prediction of fault diagnosis is based on the Self-Organizing Map (SOM). SOM is a type of artificial neural network that is trained to produce a discrete representation of the input space of the training samples, called a map. Any input may affect many parts of an aircraft. SOM separates the input for the different parts hence giving an output for each part [4].

SOM uses Multi Dimensional Scaling (MDS) in order to create its map. MDS is a set of related statistical technique often used in information visualization for exploring similarities or dissimilarities in data. MDS is a special case of ordination i.e. data clustering. It clusters similar data near each other and dissimilar data far from each other.

For high performance aircraft systems, there is a need to achieve real-time and continual assessment of aircraft condition. It is almost impossible to predict a damage of component correctly since operating conditions might be different. Hence these conditions must be accounted for.

An aircraft is affected mainly by the following factors:

1. Flying Hours
2. Aircraft Repair History
3. Airport Location
4. Climate
5. Weather

Depending on these factors (called variables in the ANN program) and compared with the OEMs, predictions are made as to the wear and tear of the parts, the next date for servicing, and replacement of any part if needed.[7]

- Any increase in temperature or altitude means a decrease in the aircraft's optimum performance. At high elevation airports, an airplane requires more runway to take off. Its rate of climb will be less, its approach will be faster, because the true air speed will be faster than the indicated air speed and the landing roll will be longer.
- Air density also decreases with temperature. Warm air is less dense than cold. As a result, on a hot day, an airplane will require more runway to take off, will have a poor rate of climb and a faster approach and will experience a longer landing roll.
- The combination of high temperature and high elevation produces a situation that aerodynamically reduces drastically the performance of the airplane. The horsepower output of the engines decrease because its fuel-air mixture is reduced. The propeller develops less thrust because the blades, as airfoils, are less efficient in the thin air. The wings develop less lift because the thin air exerts less force on the airfoils. As a result, the take-off distance is substantially increased climb performance is substantially reduced and may, in extreme situations, be non-existent.
- Depending on the number of hours an aircraft has flown in the past, and the number of repairs the aircraft has had in the past, the efficiency of the airplane changes. This calls for different treatment to planes in different conditions.

The neural network system analyzes these data of an aircraft and can hence suggest possible treatments suited to that aircraft [10]. Hence diversity can be maintained and performance can be enhanced.

## PROPOSED MRO SYSTEM

The proposed system as shown in Fig. 6 applies the concept of Artificial Neural Networks (ANN). In this system when there is any problem at any part in an aircraft then that component is reviewed. After that the OEM manuals are referred and its possible problem area and its repair are estimated and its output is fed into the ANN as an input.

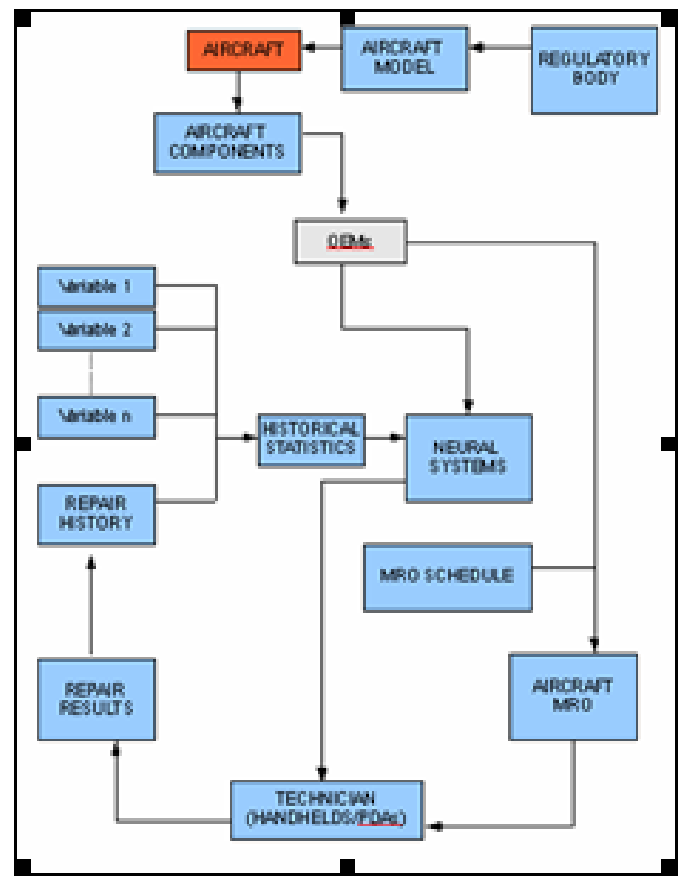

Figure 6: Block Diagram of the Proposed System

In the other half we apply the concept of Artificial Neural Network (ANN) to the system. The various factors on which the performance of an aircraft depends such as repair history, weather, climate, airports location, flying hours etc. are taken as inputs. We also need a collection of the historical data of all these variables and which will also be updated in real time as per its assessment of the result after each evaluation. These variables supported by the historical data and the input from the OEM manual are then given to the ANN. The ANN works on the basis that the high dimensional data input is converted to low dimensional with the help of Self Organizing Map (SOM) technique. The data is then clustered with the help of Multi Dimensional Scaling (MDS), in different categories in relevance to the other data inputs.[9] After that relevant logic is applied in the system to create probability prediction model of the possible percentage probability of damage in all the possible nodes of an aircraft. Many irrelevant data is eradicated by this system based on the historical data and thus the working of the system is also faster.

Now this Neural Network assesses both the input from the OEM manual and the data given by the variables and then formulates the output. The output gives the possible problem nodes in an aircraft and its repair options which is more accurate than the output given by the OEM manual. The output of the Artificial Neural Network is then fed back into the historical data as repair history which prepares the system for further reference thus building a database. This output will then be given to the technicians who will act accordingly and rectify the problem.

Advantages over the existing system:-

- Less involvement of human interface.
- Real time updation of records.
- More accurate predictions of damage.
- Helps in cost minimization of maintenance.
- Better decision making and support systems.

## CONCLUSION

The investigations carried out confirm that the usage of ANN offers rich opportunities for the diagnostics of air engines on board in real time mode. It will help detect the problem in an aircraft in a more accurate and efficient manner. Application of ANN to the system can greatly modify the creditability of the system. The survey shows that MRO spend are almost universally expected to increase, at just over 6% annually over the next five years. And if its efficiency is increased then that will help in cost minimization and better asset management for the airlines thereby increasing the performance for the aviation industry.

## REFERENCES

[1] Kohonen - books.google.com (2001).
[2] Frank Jackman - MRO Forecast
[3] John A. Bullimaria - Introduction to Neural Networks.
[4] Joseph B. Kruskal – Multi Dimensional Scaling.
[5] D.F. Garret – Aircraft System and Components.
[6] Dale Hurst – Aircraft System Maintenance.
[7] Basappa, Jategaonkar R.V. (1995) - Aspects of feed forward neural network modeling and its application to lateral-directional flight data.
[8] Ghosh A.K., Raisinghani S.C., Khubchandani S. (1998) - Estimation of aircraft lateral-directional parameters using neural networks.
[9] Hornik K., Stinchcombe M., White H. (1989) - Multi layer feed forward neural networks are universal approximators.
[10] Maine R.E., Iliff K.W. (1986) - Identification of dynamic system- application to aircraft.
[11] Jack M. Zurada – Neural Networks in Computer Intelligence.